# THE COMPLEX-VALUED ENCODING FOR DECISION-MAKING BASED ON ALIASING DATA

*P.A. Golovinski*[1,2], *V.A. Astapenko*[2]

[1,2] *Voronezh State University of Architecture and Civil Engineering, Russia, 394006, Voronezh, 20-letiya Oktyabrya street, 84*

[2] *Moscow Institute of Physics and Technology (State University), Russia, 141700, Dolgoprudnyi, Moskovskaya obl., Inststitutskii per., 9*

It is proposed a complex valued channel encoding for multidimensional data. The basic approach contains overlapping of complex nonlinear mappings. Its development leads to sparse representation of multi-channel data, increasing their dimensions and the distance between the images.

Decision-making in the control of complex systems is based on the analysis of multidimensional data. These data need to recognize them and establish functional dependences. The search for effective methods to the pattern recognition is continued. An essential step in the development of the effective methods of encoding is to create recognition systems, using sparse coding and self-learning [1]. We propose a multi-valued mapping, which improves the possibility of image classification and parameterization data for decision-making for complex multidimensional data.

The most effective encoding methods for different types of information may vary significantly. Now, the capabilities of sparse encoding are widely uses.The implementation details of the efficient numerical optimization methods are given in [2]. Often, there exists aliasing of perception in the machine learning from incomplete data. In particular, it has been studied in the considering of the reinforcement learning. This problem is also arises for the mapping, when there is no one-to-one correspondence between inputs and outputs.

In general, the perception aliasing occurs when there are two training set of data $(x_m, y_m)$, and $(x_n, y_n)$. Their inputs $x_m, x_n \in X$ are almost identical, and the expected outputs $y_m, y_n \in Y$ are different. The learning of one to one mapping $X \to Y$ in such conditions is impossible. For example, if the parameters $y_m$ and $y_n$ cause the robot to turn to the left or right, such situation would means disaster in control. To solve the problem it was proposed to build a multi-valued mapping $X \to Y$, where one input state $x_n$ corresponds to a set of outputs $y_n$ from the very beginning. A further problem of choosing between the states $y_n$ is solved in the specific training procedures.

To formulate the mapping we choose the channel representation

$$u_n = \mathrm{enc}(y_n), a_n = \mathrm{enc}(x_n), \qquad (1)$$

and train a linear map

$$u_n = Ca_n. \qquad (2)$$

Channel data processing allows represent unambiguous and multi-valued values in a unified manner [3].

The essence of the channel representation is a combination of localized nonlinear mappings with strong overlapping domains. According to Cover's theorem [4], a non-linear transformation of complex pattern into a space with higher dimension increases the probability of linear separability of sets.

Likewise, a nonlinear mapping may be used to convert the complex problem of nonlinear multiple-valued approximation to more simple linear problem. If you take a certain value, it is mapped in a collection of numbers in the channel representation, which is sparse. Most of the values therein are zero. To store the result of the mapping, one need to save non-zero values only. If we introduce two values simultaneously with different confidence scales, we get a vector mapping to larger dimension than the original was. As a result, similar initial vectors appear differently. This can be used in further classification for a response to aliased input. To encode map features, a set of complex exponential functions $a_k = \exp(i\omega(x+kT))$ is suggested. Each function is defined at the interval $[kT,(k+1)T]$ and maps the real axis to the circle. By the choice of value $\omega$ one can control a non-linear stretching of data. When the scale is $\omega = 1,$ the function recovery is based on the equation

$$\sum_{k=1}^{N} \exp[i(x+kT)]\exp(-ikT) = Ne^{ix}. \qquad (3)$$

Then

$$ix = \ln\left(\sum_{k=1}^{N} a_k \exp(-ikT)/N\right) + 2\pi n, \qquad (4)$$

where $n$ is determined by the interval $x \in [2\pi(n-1), 2\pi n]$. To calculate the matrix, connecting the inputs and outputs, the method of least squares is adopted, and

$$\min_C \|U - CA\|^2 \qquad (5)$$

with training sample $A = (a_1, a_2, \ldots a_N)$, $U = (u_1, u_2, \ldots u_N)$. As a result, the non-linear transformation to the space with large dimension allows one to adopt a linear model for relations in this new space. This approach is more naturallly realized in quantum neural networks [5].

This work was supported by RFBR (project number 11-07-00155-a).

# References

1. Olshausen B. A., Field D. J. Emergence of simple-cell receptive field properties by learning a sparse code for natural images. Nature 1996. 381. 13. P. 607-609.

2. Raina R. Self-taught learning. PhD Thesis. Ann Arbor: Stanford University, 2009. P. 1-155.

3. Granlund G.H. The complexity of vision. Signal Processing. 1999. V. 74(1). P.101-126.

4. Haykin S.O. Neural Networks and Learning Machines, NY, Prentice Hall, 2008.

5. Astapenko V.A., P.A. Golovinski. Neural networks with quantum interference. Neurocomputers: development and application. 2012. No. 4. P. 3-12.